\documentclass{article}
\usepackage{pslatex}  
\usepackage{makeidx}  
\usepackage{epsfig}  
\title{Parallel Mixed Bayesian Optimization Algorithm: A Scaleup Analysis}

\author{Jiri Ocenasek\\
Swiss Federal Institute of Technology ETH, \\
Hirschengraben 84, 8092 Z\"urich, Switzerland \\
\tt{jirio@inf.ethz.ch}\\
\\
Martin Pelikan \\
Department of Math \& Computer Science, \\
University of Missouri at St. Louis, \\
CCB 320, 8001 Natural Bridge Road, St. Louis, MO 63121 \\
\tt{pelikan@cs.umsl.edu} }

\date{}

\begin{document}

\maketitle              

\begin{abstract}
Estimation of Distribution Algorithms have been proposed as a new
paradigm for evolutionary optimization. This paper focuses on the
parallelization of Estimation of Distribution Algorithms. More
specifically, the paper discusses how to predict performance of
parallel Mixed Bayesian Optimization Algorithm (MBOA) that is
based on parallel construction of Bayesian networks with decision
trees. We determine the time complexity of parallel Mixed Bayesian
Optimization Algorithm and compare this complexity with
experimental results obtained by solving the spin glass
optimization problem. The empirical results fit well the
theoretical time complexity, so the scalability and efficiency of
parallel Mixed Bayesian Optimization Algorithm for unknown
instances of spin glass benchmarks can be predicted. Furthermore,
we derive the guidelines that can be used to design effective
parallel Estimation of Distribution Algorithms with the speedup
proportional to the number of variables in the problem.
\end{abstract}

\section{Introduction}
Estimation of distribution algorithms (EDAs)
\cite{Muehlenbein:1996,monography} , also called probabilistic
model-building algorithms (PMBGAs) \cite{survey} and iterated
density estimation evolutionary algorithms (IDEAs) \cite{idea},
are stochastic techniques that combine genetic and evolutionary
algorithms, machine learning, and statistics. EDAs are able to
effectively discover and mix the building blocks of partial
solutions, provided that the allowed complexity of probabilistic
model is relevant to optimized problem. In contrast to genetic
algorithms, their performance is not affected by the ordering of
parameters. The relations between parameters of solved problem are
discovered by machine-learning methods and incorporated into the
constructed probabilistic model.

The necessary condition of successful EDA is the generality of its
probabilistic model. The well known Bayesian Optimization
Algorithm (BOA) \cite{boa} uses a very general probabilistic model
- a Bayesian network - to encode the relations between discrete
parameters of the solved problem. Local probability distributions
in the Bayesian network in the original BOA were stored using full
conditional probability tables. More efficient representation of
the local distributions was used in the hierarchical Bayesian
Optimization Algorithm (hBOA) \cite{hBOA}, which uses decision
trees or graphs to represent these local distributions.

The Mixed Bayesian Optimization Algorithm (MBOA) \cite{mboa} is
able to deal with discrete and continuous parameters
simultaneously by using a Gaussian kernel model to capture local
distributions of continuous parameters. Its detailed principles
are described in \cite{thesis}. In binary domain it can be seen as
a variant of hBOA, but its implementation emphasizes the
possibility to efficiently perform model building in parallel. The
implementation of parallel MBOA was first simulated in the TRANSIM
tool \cite{threads} and its real implementation was reported in
\cite{Mend03}.

This paper analyzes the time complexity of parallel Mixed Bayesian
Optimization Algorithm for binary domain and compares this
complexity with experimental results obtained by solving spin
glass optimization problem. The following Section defines the spin
glass benchmark and motivates for its parallel solving. Section
\ref{Sec:MBOA} introduces the main principles of MBOA, and Section
\ref{Sec:complexity} analyzes the time complexity of each MBOA
part. Section \ref{Sec:scalability} provides the scalability
analysis of parallel MBOA, identifies the main parts that have to
be performed in parallel and derives the guidelines that can be
used to design effective parallel MBOA. The experimental results
are presented in Section \ref{Sec:empirical} and conclusions are
provided in Section \ref{Sec:conclusion}.

Note that this paper does not investigate the relation between
problem size $n$ and the minimal population size required for its
solving $N$, but it provides methods for predicting algorithmic
efficiency of parallel MBOA given the problem-dependent relation
between $n$ and $N$. See \cite{scaling} for the analysis of
population sizing in BOA.

\section{Spin glass benchmark}
\label{Sec:spinglass}
Finding the lowest energy configuration of a spin glass system is
an important task in modern quantum physics. Each configuration of
spin glass system is defined by the set of spins
\begin{equation}
S=\{s_i | \forall i \in \{1,\ldots, s^d \} : s_i \in \{+1,-1\}\},
\end{equation}
where $d$ is the dimension of spin glass grid and $s$ is the size
of spin glass grid. For the optimization by BOA the size of spin
glass problem $s^d$ is equal to the length $n$ of a solution, thus
\begin{equation}
n=s^d
\end{equation}
Each spin glass benchmark instance is defined by the set of
interactions $\{J_{i,j}\}$ between neighboring spins $s_i$ and
$s_j$ in the grid. 

The energy of the given spin glass
configuration $S$ can be computed as
\begin{equation}
E(S)=\sum\limits_{i,j \in \{1,\ldots,n\}} {J_{i,j} s_i s_j}
\end{equation}
where the sum runs over all neighboring positions $i$ and $j$ in
the grid. For general spin glass systems the interaction is a
continuous value $J_{i,j} \in < -1, +1 >$, but we focus only on
Ising model with either ferromagnetic bond $J_{i,j}=-1$ or
antiferromagnetic bond $J_{i,j}=+1$, thus $J_{i,j} \in \{-1,+1\}$.
Obviously, in the case of ferromagnetic bond the lower (negative)
contribution to the total energy is achieved if both spin are
oriented in the same direction, whereas in the case of
antiferromagnetic bond the lower (negative) contribution to the
total energy is achieved if both spins are oriented in opposite
directions. Periodic boundary conditions are used to approximate
the properties of arbitrary sized spin glass systems.

\section{Mixed Bayesian Optimization Algorithm}
\label{Sec:MBOA}
\subsection{Main principles}
The general procedure of EDA algorithm is similar to that of GA,
but the classical recombination operators are replaced by learning
and sampling a probabilistic model. First, the initial population
of EDA is generated randomly. In each generation, promising
solutions are selected from the current population of N candidate
solutions and the true probability distribution of these selected
solutions is estimated. New candidate solutions are generated by
sampling the estimated probability distribution. The new solutions
are then incorporated into the original population, replacing some
of the old solutions or all of them. The process is repeated until
the termination criteria are met.

Various models can be used in EDAs to express the underlying
probability distribution. One of the most general models is the
Bayesian network. A Bayesian network consists of two components.
The first is a directed acyclic graph (DAG) in which each vertex
corresponds to different variable of optimized problem which is
treated as random variable. This graph describes conditional
independence properties of the represented distribution. It
captures the structure of the optimized problem. The second
component is a collection of conditional probability distributions
(CPDs) that describe the conditional probability of each variable
$X_i$ given its parents ${{\Pi}}_i$ (ancestors) in the graph.
Together, these two components represent an unique probability
distribution
\begin{equation}
p(X_0 ,...,X_{n - 1} ) = \prod\limits_{i = 0}^{n - 1} {p(X_i |}
{{\Pi }}_{i} ).
\end{equation}
The well known EDA instances with Bayesian network are the
Bayesian Optimization Algorithm (BOA)\cite{boa} , the Estimation
of Bayesian Network Algorithm (EBNA) \cite{ebna} and the Learning
Factorized Distribution Algorithm (LFDA) \cite{lfda} . All these
algorithms use metrics-driven greedy algorithm for construction of
dependency graph. The implementation details of Bayesian network
construction can be found in \cite{BDe}.

\subsection{Advantages of decision trees}
The Mixed Bayesian Optimization Algorithm (MBOA) \cite{mboa} is
based on BOA with Bayesian networks, but it uses more effective
structure for representing conditional probability distributions
in the form of decision trees, as proposed in \cite{hBOA}. It
would be unreasonable to parallelize MBOA if we had no evidence
about its merits over BOA, because BOA was already parallelized in
several variants like PBOA
and DBOA \cite{dboa}.

There are several reasons for using decision trees. Firstly, only
the coefficients for those combinations of ${\Pi}_i$ that
significantly affect $X_i$ are estimated. With less coefficients
the value of each coefficient is -- on average -- estimated more
precisely. Secondly, the model building procedure allows for more
precise model, since it explores networks that would have incurred
an exponential penalty (in terms of the number of parameters
required) and thus would have not been taken into consideration
otherwise.

%
%
%
%
%
%
MBOA can be also formulated for continuous domains, where it tries
to find the partitioning of a search space into subspaces where
the parameters seem to be mutually independent. This decomposition
is captured globally by the Bayesian network model with decision
trees and the Gaussian kernel distribution is used locally to
approximate the values in each resulting partition. Without the
loss of generality, in following Sections we will focus on the
performance of MBOA in binary domain.

\section{Complexity analysis}
\label{Sec:complexity}
\subsection{Complexity of selection operator}
The most commonly used selection operator in EDAs is tournament
selection, where pairs of randomly chosen individuals compete to
take place in the parent population that serves for model
learning. The number of tournaments is $O(N)$ and for each
tournament we perform $O(n)$ steps to copy the whole chromozome,
so the total complexity is $O(n N)$. This also holds for many
other selection operators.

\subsection{Complexity of model construction}
\label{Sec:model}
%
%
The sequential hBOA builds all the decision trees at once. It
starts with empty trees and it subsequently adds decision nodes.
The quality of potential decision nodes is determined by
Bayesian-Dirichlet metrics \cite{BDe} which is able to determine
the significance of statistical correlations between combinations
of alleles in the population.
A necessary condition for adding new split node is the acyclicity
of dependency graph - it must be guaranteed that no variables
$X_i$, $X_j$ exist such that $X_i$ is used as a split in the tree
for gene $X_j$ and at the same time $X_j$ is used as a split in
the tree for gene $X_i$. In parallel MBOA it is necessary to build
each decision tree separately in different processor, so we need
an additional mechanism to guarantee the acyclicity.

In \cite{threads,Mend03} we proposed the method that solves this
problem. Each generation it uses random permutation ${\bf
o}=(o_0,o_1,…,o_{n-1})$ to predetermine topological ordering of
genes in advance. This means that only the genes can serve as
splits in the binary decision tree of target gene. The model
causality might be violated by this constraint, but from the
empirical point of view the quality of generated models is the
same as in the sequential case. The advantage is that each
processor can create the whole decision tree asynchronously and
independently of the other processors. Consequently, the linear
speedup is achieved by removing this communication overhead.
%

\medskip
\noindent
\begin{verbatim}
Function BuildTree(Population Pop, TargetVariable Xi,
             ListOfCandidateSplitVariables Pa): DecisionTreeNode;
Begin
  Initiate the frequency tables;                          ... O(n)
  For each Variable Xj in Pa                              ... O(n)
    Evaluate the metrics gain of Xj split for Xi target;  ... O(N)
  End for
  Pick up the split Xj' with the highest quality;         ... O(n)
  Pop1 := SelectIndividuals (Pop,"Xj' = 0");              ... O(N)
  Pop2 := SelectIndividuals (Pop,"Xj' = 1");              ... O(N)
  return new SplitNode(new SplitCondition("Xj'"),
                       BuildTree(Pop1,Xi,Pa-{Xj'}),
                       BuildTree(Pop2,Xi,Pa-{Xj'}));
End;
\end{verbatim}

To express the time complexity of whole model building algorithm,
we start with the complexity of one run of {\it BuildTree()}
procedure, which is $O(n+nN+n+N+N)=O(n)+O(nN)$. The total number
of {\it BuildTree()} calls is $O(2^h)$, where $h$ is the average
height of final decision tree, but note that the population size N
is decreased exponentially in the recursion:
\begin{equation}
\sum\limits_{i = 1}^h {2^h .(O(n) + O(nN/2^h )} ) \approx
\sum\limits_{i = 1}^h {2^h .O(nN/2^h )}  = \sum\limits_{i = 1}^h
{O(nN)}  = O(hnN)
\end{equation}
To be precise, the time spent on initialization of frequency
tables and the time spent on picking-up of the best split is not
compensated by this exponential decrease of population size, since
it does not depend on $N$. However, in practical cases we can
neglect this term because due to the model penalty the depth of
recursion stops at some boundary where the population size is
still reasonable such that the time required for computation of
metrics is always higher than the time for initialization of
frequency tables and for picking of the highest gain. The final
complexity of building of whole decision tree is thus $O(hnN)$ and
the complexity of building whole probabilistic model composed of
$n$ decision trees is $O(hn^2N)$. On various suites of
optimization problems we observed that the depth of decision trees
is proportional to $\log(N)$. Hence, the time complexity of model
building can be rewritten as $O(n^2 N \log(N))$. This time
complexity also holds for hBOA, but its model building algorithm
is not intended for parallel processing.

\subsection{Complexity of model sampling} Model sampling generates
$O(N)$ individuals of length $n$, where each gene is generated by
traversing down the decision tree to the leaf. On average, it
takes $O(h)=O(\log(N))$ decisions before the leaf is reached.
Thus, the overall complexity of model sampling is $O(n N
\log(N))$.

\subsection{Complexity of replacement operator}
For the MBOA algorithm we use the Restricted Tournament
Replacement (RTR) to replace a part of target population by
generated offspring. RTR was proposed in \cite{hBOA} and its code
can be specified as follows:

\medskip
\noindent
\begin{verbatim}
For each offspring individual from offspring population   ...O(N)
  For a*N randomly chosen individuals from target pop.  ...O(a*N)
    Compute the distance between chosen individual
            and offspring individual;                     ...O(n)
  End for
  If fitness of offspring individual is higher than the fitness
           of chosen individual with closest distance then
    Replace the chosen individual by offspring individual;...O(n)
  End if
End for
\end{verbatim}

The whole time complexity is then $O(N(a n N + n)) = O(a n N^2)$,
where the coefficient $a$  determines the percentage of randomly
chosen individuals in target population that undergo the
similarity comparison with each offspring individual. The greater
the $a$, the the stronger the pressure on diversity preservation.
Note that the complexity of RTR exhibits the complexity of most
other existing replacement operators.

\subsection{Complexity of fitness evaluation}
In the experimental part of this paper we use the spin glass
benchmark (defined in Section \ref{Sec:spinglass}) as a real-world
example of an NP complete optimization problem. The evaluation
time of one spin glass configuration is linearly proportional to
the number of bonds, which is linearly proportional to the number
of spins $O(n)$. For $N$ individuals the evaluation time grows
with $O(n N)$. This complexity also holds for decomposable
deceptive functions and many artificial benchmarks.

Optionally, MBOA can use hill-climbing heuristics to improve the
fitness of each individual. This heuristics tries to flip each bit
and the change with the highest fitness increase is picked and
accepted. This improvement is repeated until no flipping with
positive outcome exists. Empirically the number of successful
acceptances per chromozome is $O(\sqrt n)$ and after each
acceptance it takes $O(1)$ time to re-compute the outcomes of
neighboring flips and $O(n)$ time to pick the best flip for next
change. The total complexity of hill climbing heuristics for the
whole population is thus $O(n N \sqrt n)$. The algorithm for
picking the best flip for next change can be implemented even more
effectively, so the total complexity $O(\log(n) N n)$ can be
achieved. In our analysis we consider the case without heuristics,
but we see that the complexity of heuristics only slightly
prevails the complexity of spin glass evaluation without
heuristics.

\section{Scalability analysis}
\label{Sec:scalability}
The whole execution time of one generation
of MBOA algorithm can be formed by summing up the complexity of
selection (weighted by $c_1$), model building (weighted by $c_2$),
model sampling (weighted by $c_3$), replacement (weighted by
$c_4$), and fitness evaluation (weighted by $c_5$):

\begin{equation}
c_1 O(n N) + c_2 O(n^2 N \log(N)) + c_3 O(n N \log(N)) + c_4 O(a n
N^2 ) + c_5 O(n N)
\end{equation}

To develop scalable and efficient parallel Mixed Bayesian
Optimization Algorithm we need to identify the main tasks that are
candidates for parallelization.

Apparently, the most complex part of sequential MBOA is the
construction of probabilistic model. Also our experience with
solving spin glass benchmarks confirms this statement - typically
nearly 95\% of the execution time is spent on construction of
probabilistic model. In Section \ref{Sec:model} we outlined an
algorithm for parallel construction of Bayesian network with
decision trees proposed in \cite{threads,Mend03}. It uses the
principle of restricted ordering of nodes in graphical
probabilistic model. This method reduces the communication between
processors, so the speedup is almost linear and the model building
is able to effectively utilize up to $P=n/2$ processors.

As the first case of our analysis, we consider only this
parallelization of probabilistic model construction and keep the
other parts sequential.

With $P$ processors the overall time complexity of parallel MBOA
is

\begin{equation}
c_1 O(n N) + \frac{1}{P}c_2  O(n^2 N \log(N)) + c_3 O(n N \log(N))
+ c_4 O(a n N^2) + c_5 O(n N)
\end{equation}

Now we will analyze the proportion between sequential and parallel
part of MBOA:

\begin{equation}
\frac{{c_1 O(n N) + c_3 O(n N \log(N)) + c_4  O(a n N^2 ) + c_5
O(n N)}}{{\frac{1}{P}c_2 O(n^2 N \log(N))}} \label{eq:fraction}
\end{equation}

To obtain an algorithm that is efficiently scalable, this
proportion should approach zero as $n$ grows.

\subsection{Scalability for fixed number of processors}
We first analyze scalability in the case of constant $P$ and
increasing $n$. This is the typical scenario when the
computational resources are fixed but the problem to be solved is
very large. The detailed analysis of terms in fraction
(\ref{eq:fraction}) for $P=const.$, $n \to \infty $ gives us the
following suggestions for design of parallel MBOA:

\begin{itemize}
\item
The terms with $c_1$ and $c_5$ are negligible for scalability:
\begin{equation}
\mathop {\lim }\limits_{n \to \infty } \frac{{c_1 O(n N) + c_5 O(n
N)}}{{\frac{1}{P}c_2 O(n^2 N \log(N))}} = 0
\end{equation}
In another words, neither the selection operator nor the
population evaluation have to be implemented in parallel. Of
course this outcome is valid only for population evaluation with
time complexity $O(n N)$. For quadratic complexity of population
evaluation $O(n^2 N)$ the scalability will depend on the absolute
values of constants $c_5$, $c_2$ and on the problem-dependent
relation between "n" and "N". Theoretically, if the population
size grows linearly with problem size ($N \propto n$) it is still
possible to keep the fitness evaluation sequential, because the
$\log(N)$ term in the denominator prevails. Nevertheless, in
practical situations we suggest parallel evaluation of problems
with quadratic and higher complexity. The parallelization of
fitness evaluation is solved problem and its implementation is
straightforward. For more detailed discussion of parallel fitness
evaluation see \cite{paz}.

\item
The sampling of model does not have to be performed in parallel,
since for all possible assignments to constants $c_2$, $c_3$ and
$P$ it always holds
\begin{equation}
\mathop {\lim }\limits_{n \to \infty } \frac{{c_3 O(n N \log
(N))}}{{\frac{1}{P} c_2 O(n^2 N \log (N))}} = 0
\end{equation}

\item
The Restricted tournament replacement has not to be performed in
parallel if
\begin{equation} \mathop {\lim }\limits_{n \to \infty
} \frac{{c_4  O(a n N^2 ))}}{{\frac{1}{P} c_2 O(n^2 N \log (N))}}
= 0
\end{equation}
In this case the scalability highly depends on the
problem-dependent relation between $n$ and $N$. Theoretically,
even if the population size grows linearly with the problem size
($N \propto n$) the fraction tends to zero because the $\log(N)$
term in the denominator prevails. However, in practical situations
we suggest RTR to be performed in parallel.
\end{itemize}

\subsection{Scalability for increasing number of processors}
So far we have analyzed the scalability of sequential BOA for
fixed number of processors. Now we will analyze how the
scalability changes if the number of available processors scales
up with $n$. In this case the execution time is reduced by an
order of $n$. By assuming $P \propto n$, we obtain from Equation
(\ref{eq:fraction}):

\begin{equation}
\frac{{c_1 O(n N) + c_3 O(n N \log(N)) + c_4  O(a n N^2 ) + c_5
O(n N)}}{{c_2 O(n N \log(N))}}
\end{equation}

We see that the selection operator and the simple evaluation of
population (terms with constants $c_1$ and $c_5$) can still be
implemented sequentially, but it does not hold for fitness
evaluation with quadratic and higher complexity any more. The
decision about implementation of model sampling strongly depends
on the required speedup. If sequential model sampling is
performed, then the speedup is saturated at $c_2/c_3$. RTR has to
be necessarily implemented in parallel, because for fixed $c_4$,
$c_2$, and $a$ the numerator always prevails the denominator.

\section{Experimental results}
\label{Sec:empirical}

\subsection{Fitting complexity coefficients}
We performed a series of experiments on the random instances of 2D
Ising spin glass benchmarks of size 100, 225, 400, 625, 900 for
population sizes $N=500$, $N=1000$, $N=1500$, $N=2000$, $N=4000$,
$N=6000$ and $N=8000$. We measured separately the duration  of
each part of sequential MBOA in order to determine the
coefficients $c_1$,$c_2$,$c_3$,$c_4$,$c_5$. The fitted
coefficients are stated in Tab. \ref{coefficients}. We observed
that for larger problem sizes the fitting is in agreement with the
empirical data, but for lower problem sizes the measured time is
smaller than that expected from the theoretical complexity. This
can be explained by the effects of cache memory.

\begin{table}[h]
\begin{center}
\begin{tabular}{||c|c|c|c||}
\hline \hline
  MBOA part & Coefficient & Estimated value & R-squared value \\
\hline
  selection & $c_1$ & 8.73E-09 & $0.978$ \\
  model building & $c_2$ & 1.00E-07 & $0.979$ \\
  model sampling & $c_3$ & 1.58E-07 & $0.934$ \\
  replacement (RTR) & $c_4*a$ & 2.18E-10 & $0.989$ \\
  evaluation & $c_5$& 1.34E-07 & $0.918$ \\
\hline \hline
\end{tabular}
\newline
\caption{The resulting values of coefficients
$c_1$,$c_2$,$c_3$,$c_4$,$c_5$. For each Ising spin glass size 100,
225, 400, 625, 900 and each population size $N=500$, $N=1000$,
$N=1500$, $N=2000$, $N=4000$, $N=6000$, and $N=8000$ we choose 10
random benchmark instances and averaged the duration of each MBOA
part. The coefficients hold for one generation of MBOA performed
on Intel Pentium-4 at 2.4 GHz. The slopes of linear trend lines
were determined in MS Excel.} \label{coefficients}
\end{center}
\end{table}


\subsection{Using complexity coefficients}
The obtained coefficients can be used to predict the speedup of
MBOA with parallel model building. Given the problem size $n$, the
population size $N$, and the number of processors $P$, we get:

\begin{equation}
S = \frac{{c_1 n N + c_2 n^2 N \log(N) + c_3 n N \log(N) + c_4 a n
N^2 + c_5 n N}}{{c_1 n N + \frac{1}{P}c_2 n^2 N \log(N) + c_3 n N
\log(N) + c_4  a n N^2 + c_5 n N}} \label{eq:est_speedup}
\end{equation}

\begin{figure}
\begin{center}
\epsfig{file=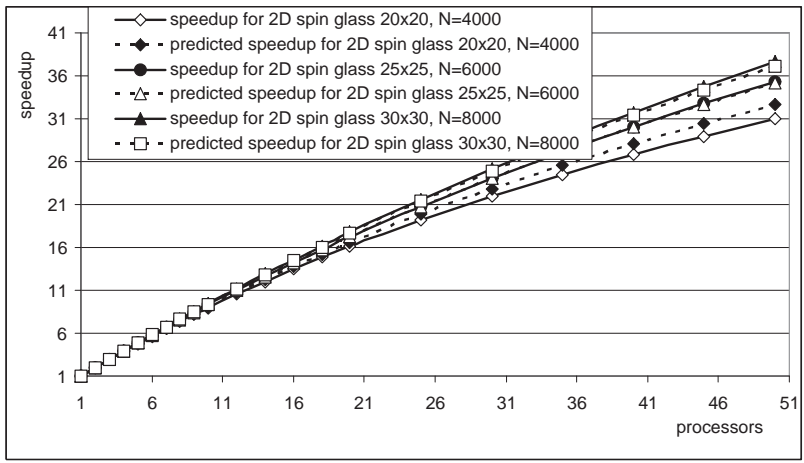, width=0.9\textwidth}
\end{center}
\caption{The comparison of the speedup predicted from the
numerical fit and the speedup computed from the empirical data
measured on sequential BOA solving 2D Ising spin glass instances
of size 20x20, 25x25, and 30x30. Population size scales
approximately linearly with the problem size. }
\label{fig:speedup}
\end{figure}

Fig. \ref{fig:speedup} shows how the predicted speedup changes for
increasing $P$ and compare it with the speedup computed from the
measured duration of each part of sequential MBOA. We considered 3
different sizes of spin glass instances 20x20, 25x25, and 30x30
and we linearly increased the population size with problem size
($N=4000$, $N=6000$, $N=8000$). As one can see, the predicted
speedup fits nicely the empirical speedup, namely for large
problem size where the caching effects disappear. Also, it can be
seen that it is possible to use larger number of processors (more
than $P=50$) without observing any significant speedup saturation.

\begin{figure}
\begin{center}
\epsfig{file=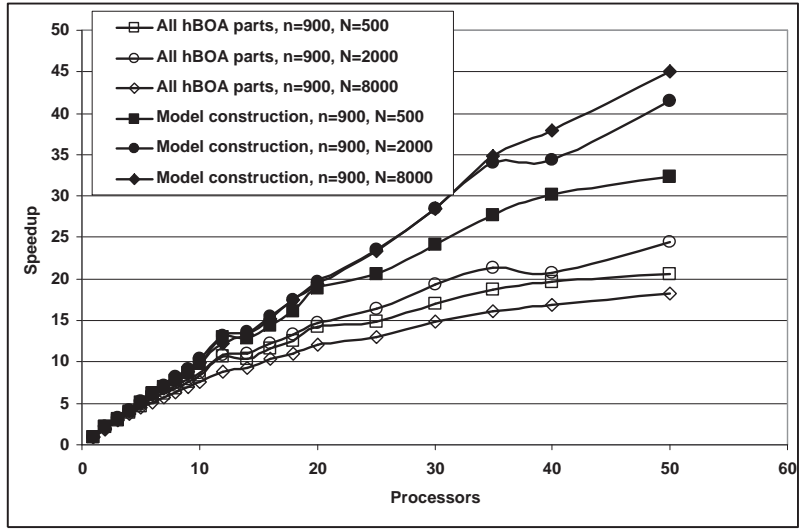, width=0.9\textwidth}
\end{center}
\caption{Speedup of the whole parallel MBOA and the speedup of
model building part (including true MPI communication) on random
spin glass instance with 30x30 spins. } \label{fig:parres}
\end{figure}

Note that Equation (\ref{eq:est_speedup}) assumes that the model
building is ideally parallelized. This is why we were able to
calculate the speedup using the time measurements from sequential
MBOA. In practical situations the communication time required by
master to communicate the population and to gather the parts of
the model from slave processors has to be considered as well. In
this analysis we neglect this term because the implementation of
message passing interface (MPI) might be platform-dependent and
its theoretical time complexity is not known. The speedup measured
on the implementation of parallel MBOA using Beowulf cluster of
502 Intel Pentium III computational nodes is shown in Fig.
\ref{fig:parres} obtained from \cite{Mend03}.

\section{Conclusions and future work}
\label{Sec:conclusion} We analyzed the time complexity of Mixed
Bayesian Optimization Algorithm and fitted this complexity to
experimental data obtained by solving spin glass optimization
problem. The empirical results fit well the theoretical time
complexity equation, so the scalability and algorithmic efficiency
of parallel Mixed Bayesian Optimization Algorithm can be
predicted, e.g. the speedup for given spin glass size and given
number of processors can be determined.

Furthermore, we derive the guidelines that can be used to design
effective parallel Estimation of Distribution Algorithms for
arbitrary optimization problems where the relation between problem
size and the minimal population size required for solving the
problem is known. Especially, we focus on the identification of
the parts of MBOA which have to be implemented in parallel.

So far the model building was the only part considered to be
performed in parallel, but our analysis identified that under some
circumstances the parallelization of other parts of MBOA might be
required as well. For example, the parallel fitness evaluation
(see \cite{paz}) can be implemented for problems with expensive
fitness evaluation. With careful parallelization of all necessary
MBOA parts the achieved speedup is proportional to the problem
size. The implementation of all critical MBOA parts in parallel
will be a subject of future work.

%
%

\end{document}